\documentclass[letterpaper]{article} 
\usepackage{aaai23}  
\usepackage{times}  
\usepackage{helvet}  
\usepackage{courier}  
\usepackage[hyphens]{url}  
\usepackage{graphicx} 
\usepackage{natbib}  
\usepackage{caption} 
\usepackage{algorithm}
\usepackage{algorithmic}

\usepackage{newfloat}
\usepackage{listings}
\DeclareCaptionStyle{ruled}{labelfont=normalfont,labelsep=colon,strut=off} 
\floatstyle{ruled}
\newfloat{listing}{tb}{lst}{}
\floatname{listing}{Listing}
\usepackage{amsmath,stmaryrd}
\usepackage{mathtools}
\usepackage[table,xcdraw]{xcolor}  
\newcommand{\Node}{\mathit{Node}}
\newcommand{\Color}{\mathit{Color}}

\title{Graphs, Constraints, and Search for the Abstraction and Reasoning Corpus}

\author{
    Yudong Xu,\textsuperscript{\rm 1}
    Elias B. Khalil,\textsuperscript{\rm 1, 2}
    Scott Sanner\textsuperscript{\rm 1}
}
\affiliations{
    \textsuperscript{\rm 1} Department of Mechanical \& Industrial Engineering, University of Toronto\\
    \textsuperscript{\rm 2} Scale AI Research Chair in Data-Driven Algorithms for Modern Supply Chains\\
    \url{wil.xu@mail.utoronto.ca}, \url{khalil@mie.utoronto.ca}, \url{ssanner@mie.utoronto.ca}
}

\usepackage{bibentry}

\begin{document}

\maketitle

\begin{abstract}
The Abstraction and Reasoning Corpus (ARC) aims at benchmarking the performance of general artificial intelligence algorithms. The ARC's focus on broad generalization and few-shot learning has made it difficult to solve using pure machine learning. A more promising approach has been to perform program synthesis within an appropriately designed Domain Specific Language (DSL). However, these too have seen limited success. We propose Abstract Reasoning with Graph Abstractions (ARGA), a new object-centric framework that first represents images using graphs and then performs a search for a correct program in a DSL that is based on the abstracted graph space. The complexity of this combinatorial search is tamed through the use of constraint acquisition, state hashing, and Tabu search. An extensive set of experiments demonstrates the promise of ARGA in tackling some of the complicated object-centric tasks of the ARC rather efficiently, producing programs that are correct and easy to understand.

\end{abstract}





\begin{figure}[t]
\centering
\includegraphics[width=1\columnwidth]{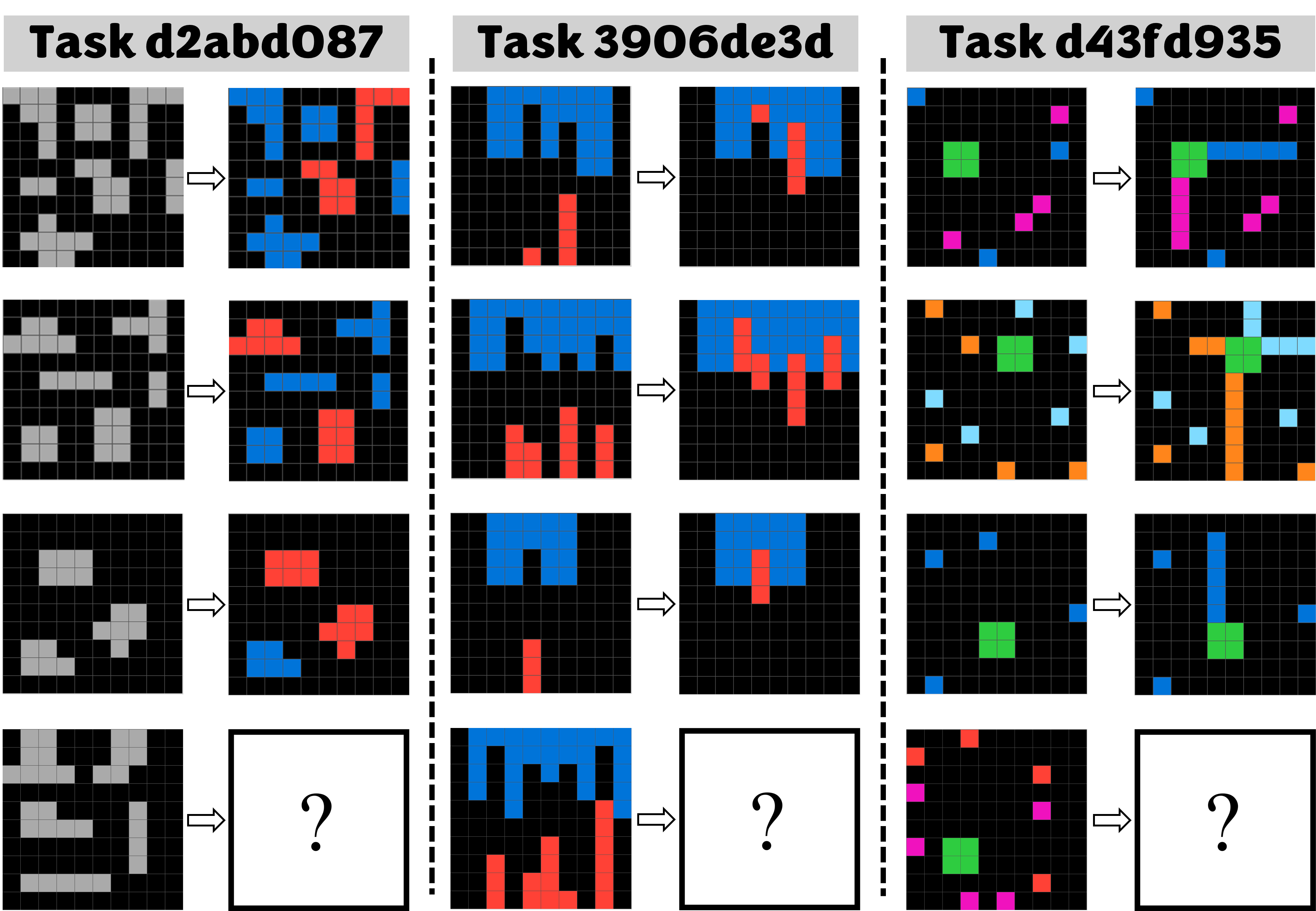} 
\caption{\textbf{Sample ARC Tasks.} Three tasks (each two consecutive columns) are shown. For a given task, each row contains one example input-output pair. The top three rows contain the ``training" instances and the bottom row contains the ``test" instance. The goal is to use the training instances to solve the test instance. The left task (``object recoloring'') requires recoloring the size-6 grey objects  to red and other grey objects to blue. The middle task (``object movement'') requires moving the red columns up until they hit the blue object. The right task (``object augmentation'') requires extending the size-1 objects directly above, below or to the sides of the green object towards it until they make contact.}
\label{fig:example-arc}
\end{figure}

In an attempt to better measure the gap between machine and human learning, the Abstraction and Reasoning Corpus (ARC) was created by~\citeauthor{c:arc} in~\citeyear{c:arc}. The dataset is a collection of 1000 image-based reasoning tasks, where each task asks for an output image given an input. To ``learn" a procedure that produces said output, each task comes with 2--5 input-output image pairs as training instances; these training inputs are different from the actual test input, but can be solved by the same (unknown) procedure. Some examples are shown in Figure \ref{fig:example-arc}. A competition with over 900 teams was hosted on Kaggle to solve the ARC~\cite{kaggle0}. Despite a massive effort, the solutions only achieved 20\% accuracy on the hidden test set, at best. In fact, the first-place solution could not solve two of the three examples shown in Figure \ref{fig:example-arc} despite their simplicity to a human.




Recognizing objects, actions performed on them, and relationships between them makes up a large portion of human cognition core systems \cite{c:core-knowledge}. The ARC embodies this notion in its tasks. In fact,~\citet{c:human-arc} found that when humans attempt to solve ARC tasks through language, half of the phrases they use relate to object detection. Therefore, an object-centric approach to solving the ARC is highly promising. Surprisingly, this key insight is yet to be leveraged.



\section{ARGA: Abstract Reasoning with Graph Abstractions}
\begin{figure*}[t]
\centering
\includegraphics[width=0.8\textwidth]{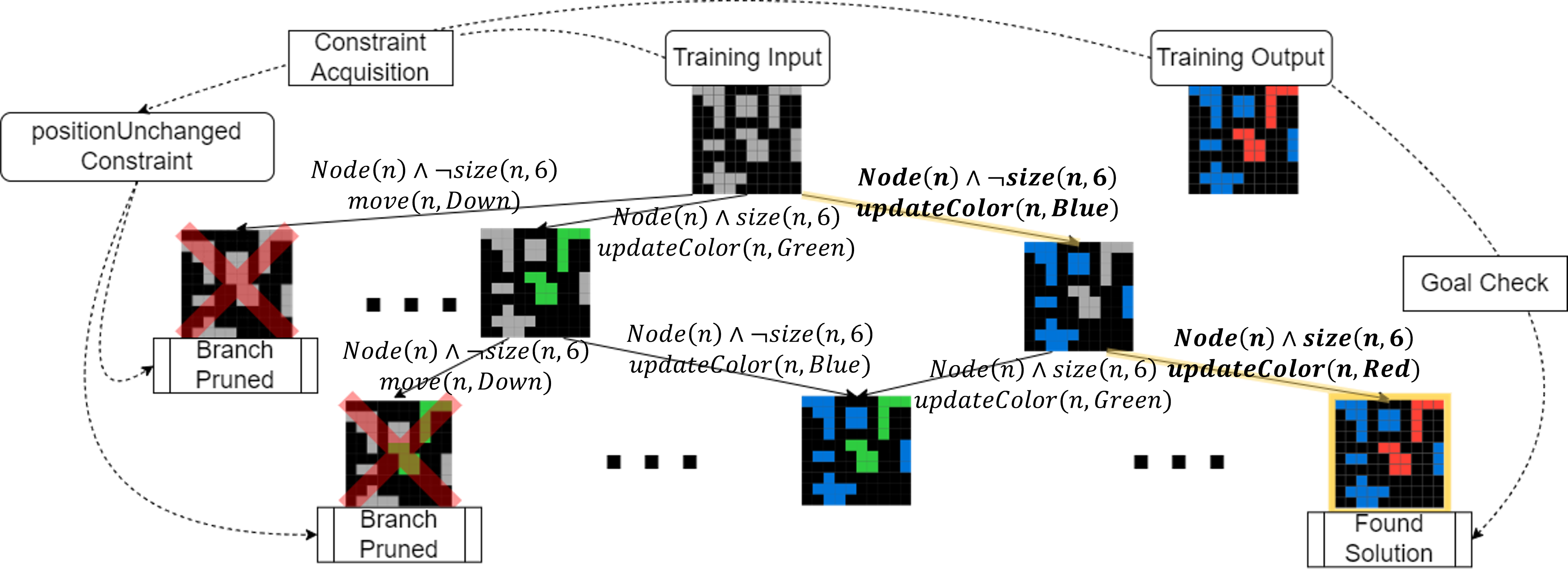} 
\caption{\textbf{Illustration of ARGA's constraint-guided search.} Note that a reconstructed 2D image is used at each node for better visualization. Nodes in the actual search tree consists a set of abstracted graphs.}
\label{fig:search-visual}
\end{figure*}

Toward this goal, we propose Abstract Reasoning with Graph Abstractions (ARGA), an object-centric framework for solving ARC tasks. 
Our design rationale is to build a computationally efficient, extensible, object-aware ARC solver through careful integration of the following:
\begin{itemize}
    \item[--] \textbf{Representation:} Enabling object awareness requires a move from treating the input as  pixels towards a \textit{graph} of objects with spatial or other relations. We design a variety of such graph abstractions to cater to the diversity of the ARC and its different definitions of objects.   
    \item[--] \textbf{Structure:} Grounded in first-order logic, our graph-based DSL makes it possible to define complex but interpretable solution programs for tasks of the ARC. This is in contrast to pure neural network-type approaches that attempt to map input to output in an often black-box fashion. 
    \item[--] \textbf{Search:} With the representation and DSL in place, we opt for a complete tree search algorithm. Given a task, the search seeks a program in the DSL that produces the correct outputs for each of the task's training examples. Whenever a correct program for a task exists in our DSL, the search can find it given sufficient time. 
    \item[--] \textbf{Constraints:} Leveraging the observation that (solved) training examples not only tell us what a correct program does but also what it should not do (e.g., in Fig.~\ref{fig:example-arc} (left), objects should not move), we use constraint acquisition to simplify our combinatorial search space. Constraints are expressed in the very same graph DSL and may be acquired by an arbitrary algorithm.

\end{itemize}
Fig.~\ref{fig:search-visual} illustrates the DSL, Search, and Constraints components of ARGA; Fig.~\ref{fig:abstraction} illustrates the Representation. With ARGA, we hope to provide AI researchers who are interested in the ARC and similar few-shot reasoning situations with the first such system upon which they can build and explore the capabilities of graph and search-based reasoning. Our implementation is available on GitHub\footnote{\url{https://github.com/khalil-research/ARGA-AAAI23}}.

Because object-oriented abstraction and reasoning are major failure modes of state-of-the-art ARC solvers, we define criteria to select a subset of object-oriented ARC tasks as a testbed for the evaluation of our methods in comparison to other top solvers. The 160 tasks in question span a wide range of challenging problems that can be categorized as object recoloring, object movement, and object augmentation.   
We show how ARGA's design and performance are favorable in the following ways:
\begin{itemize}
    \item[--] \textbf{Extensibility and Modularity:} Every component of ARGA can be extended almost independently to target additional ARC tasks or optimize performance: novel graph abstractions can be added, additional object filters and transformations can be appended to the DSL, new search strategies can be tested, and faster constraint acquisition algorithms may seamlessly replace ours. 

    \item[--] \textbf{Computational Efficiency:} Our DSL contains a number of object-based selection filters as well as transformations (e.g., recoloring, moving, etc.). Because these can be composed together to form a candidate program for an ARC task, the resulting search space is combinatorially large. Nonetheless, through experiments on 160 object-based ARC tasks, we show that when ARGA finds a solution, it does so by exploring a minute number of possible solutions, effectively three orders of magnitude fewer than the winner of the Kaggle competition. 

    \item[--] \textbf{Effectiveness:} Our current DSL includes only 4 base filters and 11 transformations. Yet, we solve 57 of 160 tasks, only slightly behind the Kaggle winner's 64 of 160. The latter includes a much larger body of transformations that were obtained by examining many more ARC tasks. 
\end{itemize}

\begin{figure*}[h]
\centering
\includegraphics[width=0.9\textwidth]{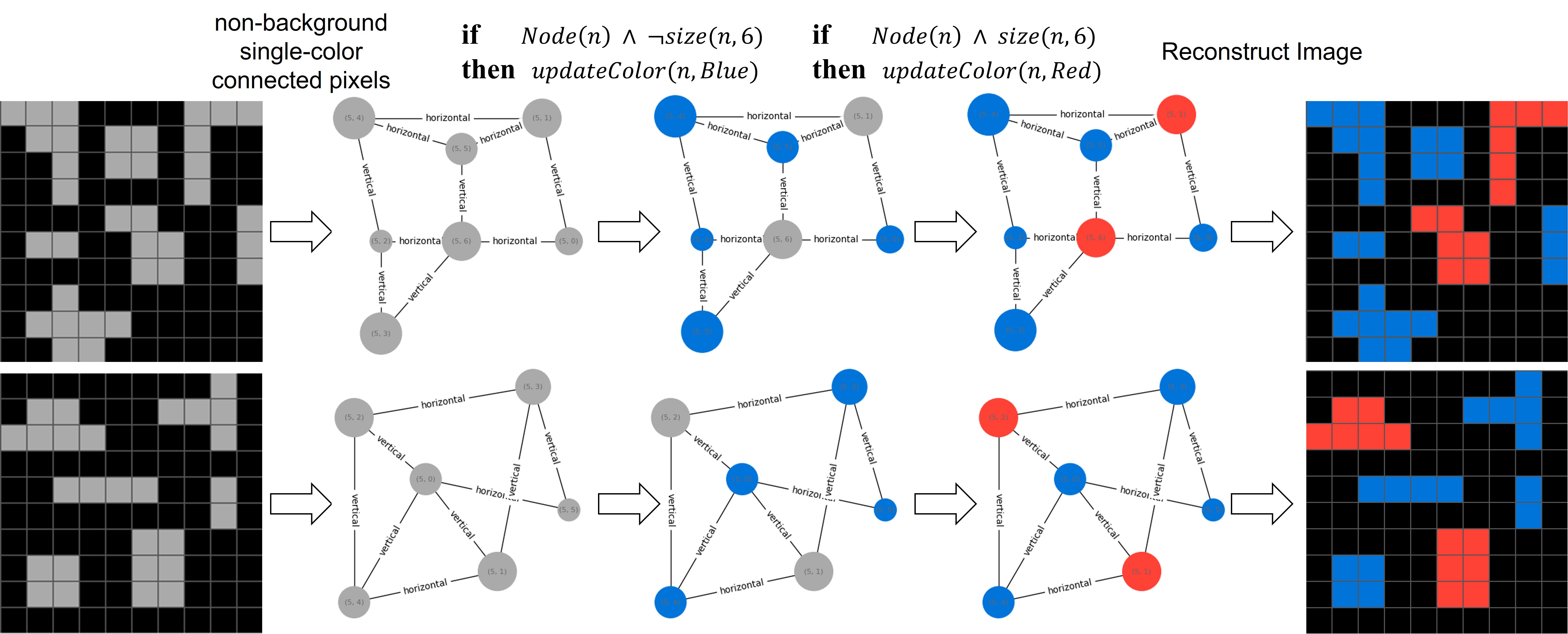} 
\caption{\textbf{Example solution generated by ARGA.} The input image is first abstracted into a graph in which each node represents a set of adjacent pixels that are not black. Two nodes share an edge iff there is at least one cell in each node with the same coordinate value along either axis. The solution here first colors in blue all nodes \textit{not} containing exactly six pixels, then colors in red all nodes with exactly six pixels. The number of pixels contained in a node is defined as its ``size'', a node attribute that ARGA can use in its search for a correct program.}
\label{fig:example-solution}
\end{figure*}

\subsection{System overview}
We propose a two-stage framework that takes an object-centric approach to solving an ARC task. First, the \emph{graph abstraction stage} inspired by work on Go~\citep{graepel2001go}, where the 2D grid images are mapped to (multiple) undirected graph representations that capture information about the objects in the images at a higher abstracted level. 
Second, the \emph{solution synthesis stage}, where a constraint-guided search is used to formulate the series of operations to be applied to the abstracted graphs that will lead to a solution. The space of possible operations is defined by an ARGA-specific relational Domain Specific Language (DSL).  

Since the DSL defines operations on the abstracted graphs, we will first describe the graph abstraction stage and formally define the structure of the abstracted graphs. Then, the DSL will be defined in detail. Finally, the solution synthesis stage will be discussed.

\section{Graph Abstraction}

Graph abstraction allows us to search for a solution at a macroscopic level. In other words, we are modifying \textit{groups of pixels} at once, instead of modifying each individual pixel separately. As a result, this approach has a smaller search space than its non-abstracted, raw image counterpart. 


We now formally introduce terminology that aids in defining our abstracted graphs (such as those shown in Figure~\ref{fig:example-solution}) that will be leveraged by the DSL of the next section. The language we use builds on first-order logic which provides a flexible and expressive language for describing typed objects and relations. Object types in our DSL are shown in Table~\ref{table:object-types} and can be used as unary predicates, e.g., $Node(n)$ is true \emph{iff} $n \in Node$.  Some example relations between objects are shown in Table~\ref{table:object-relations} and the full set of relations can be found in Appendix Table~\ref{table:object-relations-all}.


\begin{table}[t]
\centering
\begin{tabular}{ll}
\hline
Object Type Set & Object Type Description \\ \hline \hline
$i \in Image$ & A 2D grid image \\
$p \in Pixel$ & A pixel on an image \\
$g \in Graph$ & An abstracted graph \\
$n \in Node$ & A node in an abstracted graph  \\
$e \in Edge$ & An edge in an abstracted graph \\
$c \in Color$ & Color (including \emph{background})\\
$s \in Size$ & Size of a node (\# pixels)\\
$d \in Direction$ & Directions within the 2D image \\
$pa \in Pattern$ & A pattern found on the image\\
\hline
$t \in Type$ & Generic Type (any above) \\ \hline
\end{tabular}
\caption{{\bf Object Types} in ARGA.}
\label{table:object-types}
\end{table}

\begin{table}[t]
\resizebox{\columnwidth}{!}{
\begin{tabular}{ll}
\hline
Typed Object Binary Relations & Description \\ \hline \hline
$containsNode(Graph,Node)$ & Graph contains Node  \\
$containsPixel(Node,Pixel)$ & Node contains Pixel\\
$neighbor(Node,Node)$ & An edge exists between two Nodes \\
$color(Node,Color)$ & color of Node \\ 
$size(Node,Size)$ & size of Node \\ 
\hline
$Rel(Type,Type)$ & Generic Relation (any above) \\
\hline
\end{tabular}}
\caption{\textbf{Example Object Relations} in ARGA.}
\label{table:object-relations}
\end{table}

\begin{figure}
\centering
\includegraphics[width=0.9\columnwidth]{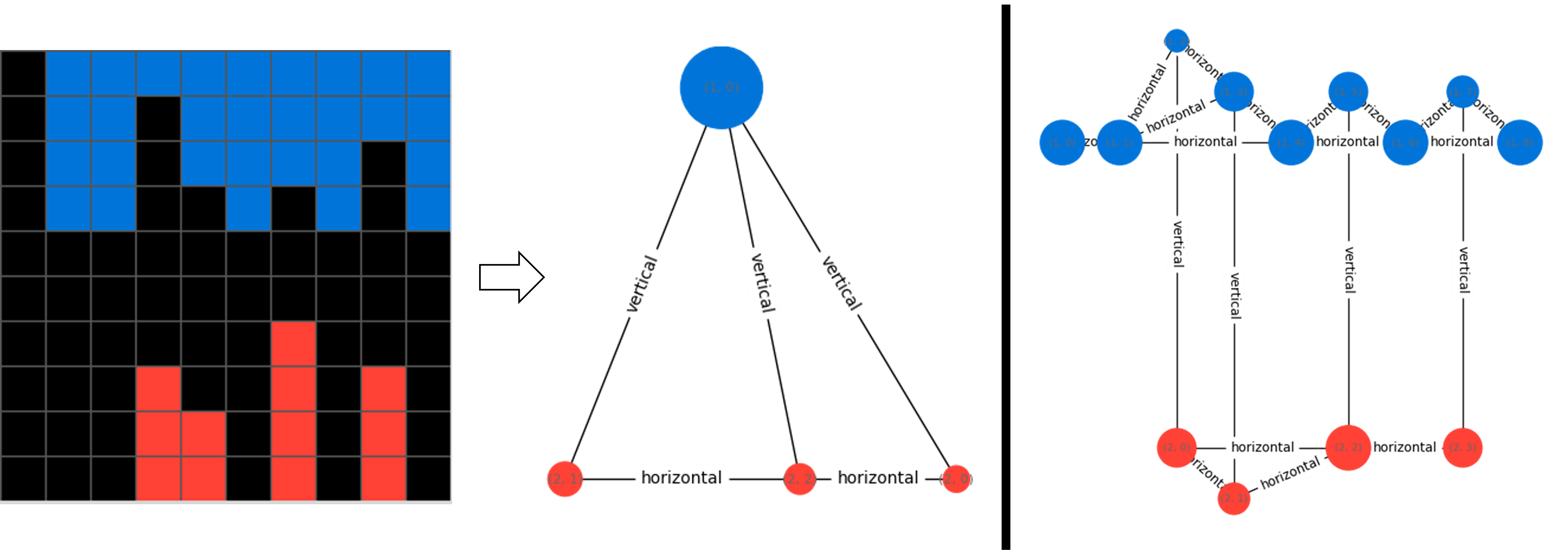} 
\caption{\textbf{Visualization of graph abstractions.} Applying two different graph abstractions to an image. Left: non-background single-color connected pixels. Right: non-background single-color vertically-connected pixels.}
\label{fig:abstraction}
\end{figure}

Let $i$ be any input or output 2D grid image from an ARC task. $i$ can be completely specified by its set of pixels $p$. Let $g$ be an abstracted graph with sets of abstracted nodes $n$. 
The relations that hold between these types are shown in Table \ref{table:object-relations}. 
Each Node $n$ represents an object that is detected in the original image $i$ based on the rules of the abstraction (e.g., one graph abstraction is ``non-black neighboring pixels of the same color form a node'') and relations between the nodes represent relationships between these objects. 

Therefore, the graph abstraction process executes a mapping that generates some abstracted graph $G$ for image $I$. We note that there are multiple ways in which this mapping can be defined. Different graph abstractions can be used to identify objects in the image using different definitions of what an object is.
Since the resulting abstracted graphs from different graph abstraction definitions share the same underlying structure, we are able to expand the solution space significantly without modifying the DSL. 

The usefulness of having multiple definitions of an object can be observed in the example shown in Figure~\ref{fig:example-arc} (Middle). Upon first inspection, one may think that objects are defined as connected pixels with the same color. However, upon further inspection, we realize that the connected red pixels in different columns are in fact different objects as they do not share the same modification in the output images. Therefore, defining multiple abstraction processes improves ARGA's ability to correctly capture object information from the input images. The two different abstractions mentioned in the example are further discussed in the following: 

\noindent\textbf{non-background single-color connected pixels:} 
In this abstraction, an object (or node) is defined as a set of connected pixels sharing the same color. The pseudocode for the abstraction algorithm is shown in Appendix Algorithm \ref{alg:abstract} and an illustration of this abstraction is in Figure~\ref{fig:abstraction}.
%

\noindent\textbf{non-background single-color vertically-connected pixels:}
In this graph abstraction, an object is defined as a set of vertically connected pixels that are not the background color. An illustration of this abstraction is in Figure~\ref{fig:abstraction}.


\subsection{Overlapping Objects}
Note that our representation allows for a pixel to be associated with multiple nodes in the graph, as objects are modified. This can be intuitively understood by observing that objects may overlap with one another on the grid as one applies a sequence of transformations to solve a given task. Our graph abstraction ensures that although some objects may be partially obscured, they are still kept track of and considered to be a whole object. This allows the system to have the \textit{object persistence} knowledge prior.

\section{A Graph DSL for the ARC}



We now introduce a lifted relational DSL for ARGA built upon the objects and relations defined in the previous section. The DSL is used to formally describe the filter language used to match node patterns, determine graph transformation parameters, and carry out transformations on abstracted graphs as described in the following. An example solution expressed using the DSL is shown in Figure~\ref{fig:example-solution}.

\subsection{Filters}

Filters are used to select nodes from the graph. The fundamental \emph{grammar} is a subset of first-order logic:
\begin{align*}
Filter(x) & \Coloneqq Type(x) \\
& \Coloneqq Filter(x) \land Filter(x) \\
& \Coloneqq Filter(x) \lor Filter(x) \\
& \Coloneqq \neg Filter(x) \\
& \Coloneqq \exists{y} \, Rel(x,y) \land Filter(y)\\
& \Coloneqq \exists{y} \, Rel(y,x) \land Filter(y)\\
& \Coloneqq \forall{y} \, Rel(x,y) \implies Filter(y) \\
& \Coloneqq \forall{y} \, Rel(y,x) \implies  Filter(y)\\
& \Coloneqq Rel(x,c) \; \textrm{[$c$ is a constant]} \\
& \Coloneqq Rel(c,x) \; \textrm{[$c$ is a constant]}
\end{align*}

The following example filters match nodes with 6 pixels, with grey as their color, and whose neighbors are all blue, respectively:
\begin{align*}
\mathit{filterBySize6}(n) \equiv & \, \Node(n) \land size(n, 6)\\
\mathit{filterByColorGrey}(n) \equiv & \, \Node(n) \land color(n,grey) \\
\mathit{neighborsAllBlue}(n) \equiv & \, \Node(n) \\
 \land \forall{y} \, \mathit{Neighbor}(n,y) &  \implies color(y,blue)
\end{align*}




\subsection{Transformations}



Transformations are used to modify nodes selected by filters. They do so by modifying the values of object relations. Table~\ref{table:transformations} describes a few of the transformations; the full list can be found in Appendix Table~\ref{table:transformationsfull}. 

\begin{table}[h!]
\centering
\resizebox{\columnwidth}{!}{%
\begin{tabular}{lp{0.55\columnwidth}}
\hline
Transformation & Description \\ \hline \hline
$\mathit{updateColor}(\Node, Color)$ & Update color of Node to Color\\
$\mathit{move}(\Node, \mathit{Direction})$ & Update pixels of Node to move in Direction\\
$\mathit{rotate}(\Node)$ & Update pixels of N to rotate it clockwise\\
$\mathit{extend}(\Node, \mathit{Direction})$ & Add additional pixels to Node in Direction\\
\hline
$\mathit{transform}(\Node, v_1,\ldots,v_k)$ & Generic transformation with \mbox{k parameter values $(v_1,\ldots,v_k)$}\\
\hline
\end{tabular}}
\caption{{\bf Example Transformations}.}
\label{table:transformations}
\end{table}

An example transformation definition is shown below. 
\begin{align*}
& updateColor (n:\Node, c:\Color) \\
& \longrightarrow color(n,c) \wedge \neg color(n,c') \quad \forall c' \in \Color \textrm{ s.t. } c' \neq c
\end{align*}
In this example, the transformation \textit{updateColor} updates ($\longrightarrow$) the color of the Node $n$ to $c$. It does so by assigning $color(n,c)$ to true and $color(n,c')$ to false for all other colors $c'$ in the abstracted graph representation.  


\subsection{Dynamic Parameter Transformations}

\begin{figure}[t]
\centering
\includegraphics[width=1\columnwidth]{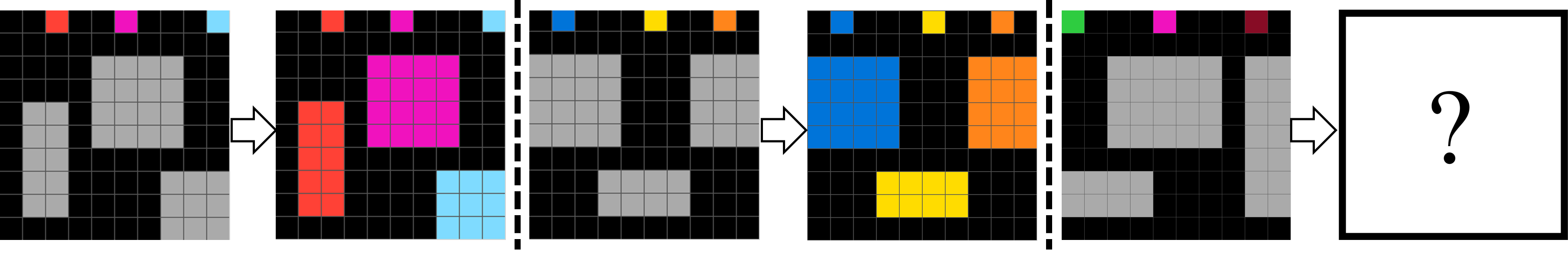} 
\caption{\textbf{Example Task from the ARC that requires dynamic transformation parameter.} The target color of a grey node is determined dynamically based on the input.}
\label{fig:example-dynamic}
\end{figure}

In the example shown in Figure \ref{fig:example-arc} (Left), we can ``statically" identify the color that the nodes should be updated to. However, this does not work for Figure \ref{fig:example-dynamic}, because the target color of a transformed grey object is that of its neighboring size-1 object. Therefore, we define parameter binding functions which allow us to dynamically generate parameters for transformations. The grammar for parameter binding as well as its interpretation and an example are provided next: 
\begin{align*}
Param(x,v) \Coloneqq & v = c \, \textrm{[$c$ is a constant]}\\
\Coloneqq & Rel(x,v) \\
\Coloneqq & Rel(v,x) \\
\Coloneqq & \exists{y} \, Rel(x,y) \land Filter(y) \land Param(y,v)\\
\Coloneqq & \exists{y} \, Rel(y,x) \land Filter(y) \land Param(y,v)
\end{align*}
While it shares a similar grammar to filters, the $Param(x,v)$ has special semantics that we pause to discuss.  First, the goal of $Param(x,v)$ is to find possible matching parameters for an object $x$, hence we never apply a filter to $x$ in the grammar since we are not aiming to restrict it --- $x$ is assumed to be given.  Second, we can interpret $Param(x,v)$ as providing all possible parameter values $v$ that make $Param(x,v)$ true.  However, we need a unique parameter $v$; if no $v$ matches for a given $x$ then $Param(x,v)$ fails to return a parameter and we cannot apply the transformation (it is considered a \emph{noop}).  If multiple $v$ match, then we deterministically order and return the first matching $v$.  While this is generally undesirable behavior, we note that our search over dynamic parameter bindings $Param(x,v)$ most often only succeeds when $Param(x,v)$ represents a \emph{functional} matching such as $Param(x,v) \equiv Color(x,v)$ since we know that Color is an injective relation.  Hence, we do not \emph{a priori} restrict the grammar search to functional parameter bindings, but find in practice that successful $Param(x,v)$ bindings found in search tend to recover functional mappings from $x \mapsto v$ based on invariant properties inherent in the training examples.

We remark that this dynamic parameter grammar includes static cases such as $Param(x,v) \equiv v=blue$, which would ignore the node $x$ and always return the parameter $blue$.

Following is a more complex parameter binding: 
\begin{align*}
bindSize1NeighborColor&(x,v) \equiv \\
\exists{y} \, neighbor(x,y) & \land size(y,1) \land Color(y, v))
\end{align*}
Here, $bindSize1NeighborColor(x,v)$ matches (and returns) the color $v$ of any neighbor of $x$ with a size of 1 pixel.
In the example shown in Figure~\ref{fig:example-dynamic}, suppose we have grey Node \textit{n} selected by filters; we can then find the color to update it by calling $bindSize1NeighborColor(n,Color)$.

\subsection{Full Operation}

With the filters, transformations and parameter bindings formally defined, we may now combine them to perform a full modification to the abstracted graph. Given a filter, a transformation, and $k$ parameter bindings $Param_i(x,v)$ ($i \in \{1 \ldots k\}$) for each parameter taken by the transformation (possibly none if $k=0$): 
\begin{align*}
\text{for each} \quad & n \in Node \\
\text{if} \quad & filter(n) \\
\text{then} \quad & v_i \mapsfrom \{ v | Param_i(n,v) \} \text{ for } i \in \{ 1\ldots k\} \\
& transform(n, v_1, \ldots, v_k)
\end{align*}
We assume that $\mapsfrom$ deterministically selects a unique value $v_i$ if $|\{v\}| \neq 1$.
The set of operations required for solving the example in Figure~\ref{fig:example-dynamic} are $filterByColorGrey$, $updateColor$ and $bindSize1NeighborColor$.
We note that tasks such as the example shown in Figure \ref{fig:example-solution} do not require dynamic parameters; in those instances, the parameter binding found in the solution simply returns a static value $v = c$.

\section{Solution Synthesis}

With a DSL clearly defining the solution space and the input images successfully abstracted, a search will be conducted to synthesize a solution. 
Many ARC tasks have very complicated logic with multiple detectable objects, which means that even with our high-level graph abstraction, the search space is too large to be explored exhaustively. Therefore, the key objective in developing our algorithm is to reduce the search space. To achieve this, we introduce a \textit{constraint acquisition} module which acquires constraints that are used to prune unpromising branches of the search tree, i.e., sequences of transformations which cannot possibly result in a correct solution to the training tasks. Other tricks such as hashing and Tabu List are also used to speed up the search. An illustration of a search tree is shown in Figure \ref{fig:search-visual}

\subsection{Search Strategy}
We implement a greedy best-first search. Suppose ARC task $t$ has $m$ training instances, with input-output images $\{ \mathit{input_i}, \mathit{output_i} \} \text{ for } i \in \{1,\dots, m\}$. Each node in our search tree contains a set of graphs $\{ g_{input\_i} \} \text{ for } i \in \{1,\dots, m\}$. $g_{input\_i}$ represents $input_i$ after the abstraction process and the application of a sequence of operations $(o_1,\dots,o_j,\dots,o_k) \text{ for } j \in \{1,\dots, k\}$, where each $o_j$ is a full operation as defined previously. The special case of $k=0$  corresponds to no operations applied, i.e., the root node of the search tree. 

To expand a node with abstracted graphs $\{ g_{input\_i} \}$, we first identify the set of all valid full operations $O$. Then, for each $ o \in O $, we apply it to $\{ g_{input\_i} \}$ and obtain updated abstracted graphs $\{ g'_i \} \text{ for } i \in \{1,\dots, m\}$. We add the new abstracted graphs $\{ g'_i \}$ into the search tree as a new node and update the sequence of operations that led to it by appending operation $o$ to obtain $(o_1,\dots,o_k,o)$. 

\subsection{Heuristic Function}
To determine the node to be expanded in each iteration of the search, our primary metric measures how close the node is to the target training output. For each node, we reconstruct the corresponding 2D image for each of the abstracted graphs $\{ g_{input\_i} \}$. We then compare the reconstructed images with the training outputs $\{ output_i \}$ and calculate a penalty score based on pixel-wise accuracy, as detailed in Appendix Table~\ref{table:cost-function}. Large mismatch in pixels between the ``predicted" and the actual output results in a large penalty. The node in the search tree with the lowest score is selected for expansion. 
 
\subsection{Constraint-Guided Search}


We illustrate this concept with an example. All objects in Figure \ref{fig:example-arc} (Left) should not change in position. We can therefore define the constraint \textit{positionUnchanged}, which is satisfied when a node and the updated version of that node share the same set of pixels, thus making sure that the node's position on the image remains unchanged through the transformation. All transformations that modify a node's pixels can therefore be pruned by this constraint in the search tree. A visualization is shown in Figure \ref{fig:search-visual}.

The constraints can be defined using the same language we've introduced earlier. For instance, \textit{positionUnchanged} is defined as:
\begin{align*}
& positionUnchanged (n:Node, n':Node)\equiv\\
& \forall p \in P \enspace containsPixel(n,p) \equiv  containsPixel(n',p)
\end{align*}
which holds if for all pixels $p \in P$, $\mathit{containsPixel} (n,p)$ and $\mathit{containsPixel} (n',p)$ return the same value. Constraints defined for ARGA can be found in Appendix Table~\ref{table:constraints}.


Given a set of constraints $C$ that must be satisfied and a node in the search tree with a set of graphs $\{ g_{input\_i} \} \text{ for } i \in \{1,\dots, m\}$, the search space is pruned as follows. Suppose we have a full operation $o$ that selects $n$ from $g_{input\_i}$ with filter operation $f$  and transforms it with operation $t$ to produce updated node $n'$. If $\exists c \in C \wedge c(n, n') = False$, then the branch in the search tree created by applying $o$ to $\{g_{input\_i}\}$ is pruned, as it does not satisfy constraint $c$.


\subsection{Constraint Acquisition}

To obtain a set of constraints to prune the search space, we introduce a simple constraint acquisition algorithm inspired by the ModelSeeker~\cite{model-seeker} and Inductive Logic Programming \cite{ca-ilp}. 

We have the generic constraint $\mathit{constraint}(n:\Node,n':\Node)$ where $n, n'$ can be understood as a node before and after modification by a transformation. To determine the constraints that must hold for a particular ARC task from the set of all possible constraints, we compare the training output images to the corresponding input images. 

While expanding a node in the search tree, we apply the same abstraction process for the output images $\{output_i\} $ as the input images to obtain  $\{ g_{output\_i} \} \text{ for } i \in \{1,\dots, m\}$. Then, for each full operation $ o \in O $, we apply its filter operation $f$ to $ g_{input\_i} \text{ and } g_{output\_i}$ to obtain pairs of $n_{in}$ and $n_{out}$.  For each constraint $c$, if $c(n_{in}, n_{out})$ evaluates to True for all pairs found by $f$, we say that constraint $c$ must be satisfied for all nodes selected by filter $f$. Therefore, all full operations $o$ with filter $f$ and transformation $t$ that violate constraint $c$ can be pruned.

\subsection{Hashing}

It is highly likely that different transformations or sequences of transformations result in the same abstracted graph. To avoid duplicate search efforts, we hash each node in the search tree so that equivalent nodes are only explored once. The search tree therefore has the structure of a Directed Acyclic Graph. An example of this is shown in Figure \ref{fig:search-visual}.

\subsection{Tabu List}

In our current implementation, abstracted graphs from different abstractions share the same search tree. It is therefore possible that greedy best-first search will get stuck in unpromising local solutions. To avoid this, we implement a simple Tabu List, which keeps tracks of the performance of each abstraction. If an abstraction is generating increasingly worse results, we temporarily place it on the Tabu List so that no nodes with this abstraction will be explored. 

 \begin{table*}[htbp!]
\centering
\begin{tabular}{llrrll}
\hline
 Model & Task Type    & \# Training Correct & \# Testing Correct & Average Nodes & Average Time (sec.) \\ \hline \hline
 ARGA &movement     & 18/31 (58.06\%)    & 17/31 (54.84\%)   & 3830.35                              & 89.75                    \\
 &recolor      & 25/62 (40.32\%)    & 23/62 (37.10\%)   & 12316.87                             & 326.83                   \\
 &augmentation & 20/67 (29.85\%)    & 17/67 (25.37\%)   & 4668.82                              & 67.09                    \\
 &all          & 63/160 (39.38\%)   & 57/160 (35.62\%)  & 7504.81                              & 178.66                   \\
 \hline
 Kaggle &movement     & 21/31 (67.74\%)    & 15/31 (48.39\%)   & 2176777.67                           & 62.45                    \\
 First Place&recolor      & 23/62 (37.10\%)    & 28/62 (45.16\%)   & 2290441.32                           & 93.19                    \\
 &augmentation & 35/67 (52.24\%)    & 21/67 (31.34\%)   & 2248151.10                           & 66.07                    \\
 &all          & 79/160 (49.38\%)   & 64/160 (40.00\%)  & 2249924.92                           & 77.08                    \\
\hline
\end{tabular}
\caption{\textbf{Results on subset of ARC.} \textit{\# Training correct} is the number of tasks that got all the training instances exactly right. \textit{\# Testing correct} is the number of tasks that got the testing instance exactly right. \textit{Average Nodes} is the average number of unique nodes added to the search tree before finding a solution for correctly solved tasks. \textit{Average Time (sec.)} is the average time in seconds to reach solution for correctly solved tasks.}
\label{tab:results}
\end{table*}

\begin{table}[htbp!]
\centering
\resizebox{\columnwidth}{!}{
\begin{tabular}{lccp{0.2\linewidth}p{0.2\linewidth}}
\hline
 Model                 & \# Training Correct & \# Testing Correct & Average Nodes & Average Time (sec.)\\ \hline \hline
ARGA               & 63 (39.38\%)   & 57 (35.62\%)  & 7504.81                              & 178.66                   \\
-CA     & 62 (38.75\%)   & 55 (34.38\%)  & 12114.25                             & 227.62                   \\
-SF & 60 (37.50\%)   & 54 (33.75\%)  & 8530.17                              & 197.54                   \\
-TL & 64 (40.00\%)   & 57 (35.62\%)  & 7702.53                              & 169.52                   \\
-H & 62 (38.75\%)   & 57 (35.62\%)  & 26107.58                             & 172.77 \\
\hline
\end{tabular}
}
\caption{\textbf{Ablation study.} ARGA is the complete system. -CA is ARGA without constraint acquisition. -SF is ARGA using a breadth-first search strategy for  abstractions. -TL is ARGA without Tabu List. -H is ARGA without hashing.}
\label{tab:ablation}
\end{table}

\section{Experiments}

\citet{c:arc} states that the ARC aims to evaluate ``Developer-aware generalization'', and all ARC tasks are unique and do not assume any knowledge other than the core priors. Therefore, implementing and evaluating ARGA on a subset of ARC tasks should provide useful insight into the effectiveness of our method without the need for extensive development of transformation functions, which are not the focus of our contribution. 

We focus on a subset of 160 object-centric tasks from the ARC and categorize them into three groups: (1) \emph{Object Recoloring} tasks, which change colors of some objects in the input image; (2) \emph{Object Movement} tasks, which change the position of some objects in the input image; (3) \emph{Object Augmentation} tasks, which expand or add certain patterns to objects from the input images. An example task from each of the three sub-categories is shown in Figure \ref{fig:example-arc}. 

For comparison, we evaluated the Kaggle Challenge's first-place model~\cite{kaggle1} on the same subset of tasks. The model was executed without the time limitation enforced by the competition and the highest-scored candidate produced by the model was used to generate the final prediction. 


\subsection{Results}

The performance of ARGA and the Kaggle competition's first-place solution are shown in Table \ref{tab:results}. With the exception of Object Movement tasks, our model performed slightly worse than the Kaggle winner in terms of accuracy. This is likely due to the solution space spanned by our DSL not being expressive enough, as it was developed using only a subset of the 160 tasks. On the other hand, the DSL used in the Kaggle solution was developed by first manually solving 200 tasks from the ARC \cite{kaggle1}.

Despite lower accuracy, ARGA achieves much better efficiency in search as we are able to reach the solution with 3 order magnitude fewer nodes explored. This suggests that with a more expressive DSL and a more efficient implementation, ARGA should be able to solve more tasks with much less search effort (ARGA is currently implemented in Python while the Kaggle solution is implemented in C++).


Furthermore, the gap between the number of tasks for which all training instances are solved (\# Training Correct) and the number of tasks for which the single test instance is solved (\# Testing Correct) is much smaller for ARGA. This suggests that ARGA is better at finding solutions which generalize correctly while the Kaggle solution often overfits to the training instances.




\subsubsection{Ablation Study}

Table \ref{tab:ablation} shows the performance of different variations of ARGA; the accuracies are reported on all 160 tasks. We see that the use of constraint acquisition is very effective in reducing the search space, resulting in 38\% lower average nodes explored before reaching the solution. Furthermore, the results show that Tabu List, hashing, as well as the proper searching strategy are all important for the best performance. We note that as seen in Appendix Table~\ref{table:solvedoverlaps}, there are no significant differences in the sets of tasks solved by variations of ARGA.   

\section{Related Work}


\subsubsection{Current ARC Solvers}
There have been many attempts at solving the ARC. Most of those that have shown some success leverage a DSL within the program synthesis paradigm \cite{kaggle0}. It has been shown that humans are able to compose a set of natural language instructions that are expressive enough to solve most of the ARC tasks, which suggests that the ARC is solvable with a powerful enough DSL and an efficient program synthesis algorithm \cite{c:human-arc2}. 
Indeed, this is the approach suggested by \citet{c:arc} when introducing the dataset: ``A hypothetical ARC solver may take the form of a program synthesis engine'' that ``generate candidates that transform input grids into output grids.'' 

Solutions using this approach include the winner of the Kaggle challenge, where the DSL was created by manually solving ARC tasks and the program synthesis algorithm is a search that utilizes directed acyclic graphs (DAG). Each node in the DAG is an image, and edges between the nodes are transformations \cite{kaggle1}. The second-place solution introduces a preprocessing stage before following a similar approach \cite{kaggle2}. Many other Kaggle top performers share this approach \cite{kaggle3, kaggle4, kaggle5}. \citet{grammar-arc} propose a Grammatical Evolution algorithm to generate solutions within their DSL. \citet{dreamcoderarc} utilize an existing program synthesis system called DreamCoder~\cite{dreamcoder} to create abstractions from a simple DSL through the process of compression. The program then composes the solution for new tasks using neural-guided synthesis.

Other approaches to solving the ARC include the Neural Abstract Reasoner, which is a deep learning method that succeeds in a subset of the ARC’s problems \cite{nar}. \citet{object-arc} developed a compositional imagination approach which generates unseen tasks for better generalization. \citet{discriptivegrid} develops an approach based on descriptive grids. However, these approaches have not achieved state-of-the-art results. 









\subsubsection{Constraint Acquisition}

 (CA) is a field that aims to generate Constraint Programming (CP) models from examples \cite{ca_overview}. State-of-the-art CA algorithms may be active, requiring interaction from the user~\cite{quacq, multiacq}, or passive, requiring only initial examples~\cite{conacqold}.

The passive CA algorithm used for ARGA was influenced by ModelSeeker \cite{model-seeker}, which finds relevant constraints from the global constraint catalog \cite{global-constraint-catalog} as well as the system developed by \citet{ca-ilp} which uses Inductive Logic Programming (ILP) and formulates constraints from logical interpretations. 


\section{Conclusion}
We proposed Abstract Reasoning with Graph Abstractions (ARGA), an object-centric framework that solves ARC tasks by first generating graph abstractions and then performing a constraint-guided search. We evaluated our framework on an object-centric subset of the ARC dataset and obtained promising results. Notably, the efficiency in reaching the solution within the search space shows that with further development of the DSL, our method has the potential to solve far more complicated problems than state-of-the-art methods.

\bibliography{aaai23.bib}

\clearpage
\appendix

\section{Technical Details}
\begin{algorithm}[h]
\textbf{Input}: Grid Image \textit{I}\\
\textbf{Output}: Abstracted Graph \textit{G}
\begin{algorithmic}[1] 
\STATE Identify background color \textit{background-color}
\STATE Construct 2D grid graph \textit{I'} for image \textit{I} with node for each pixel in the image and edge between each adjacent pixel. 
\STATE Initialize abstracted graph \textit{G}
\FOR{\textit{color} in all available non-background color}
\STATE Find sub-graph \textit{SG} of \textit{I'} with \textit{node.Color == color}
\STATE Find connected components \textit{C} of \textit{SG}
\FOR{component in \textit{C}}
\STATE Add node to \textit{G}  
\STATE G.pixels = component
\STATE G.color = color
\ENDFOR
\ENDFOR
\STATE Add edges between nodes based on image \textit{I} with relation vertical or horizontal
\STATE \textbf{return} \textit{G}
\end{algorithmic}
\caption{non-background single-color connected pixels graph abstraction}
\label{alg:abstract}
\end{algorithm}

\begin{table}[h]
\resizebox{\columnwidth}{!}{
\begin{tabular}{ll}
\hline
Typed Object Binary Relations & Description \\ \hline \hline
$containsNode(Graph,Node)$ & Graph contains Node  \\
$containsPixel(Node,Pixel)$ & Node contains Pixel\\
$edgeSource(Node,Edge)$ & Node is the source node for Edge\\
$edgeTarget(Edge,Node)$ & Node is the target node for Edge \\
$direction(Edge,Direction)$ & direction of Edge \\
$overlap(Node,Node)$ & Two Nodes are overlapping \\
$neighbor(Node,Node)$ & An edge exists between two Nodes \\
$color(Node,Color)$ & color of Node \\ 
$size(Node,Size)$ & size of Node \\ 
\hline
$Rel(Type,Type)$ & Generic Relation (any above) \\
\hline
\end{tabular}}
\caption{\textbf{Full List of Object Relations} in ARGA. Further quantitative information can be introduced by implementing new relations. For example, to account for distance between two nodes, we can introduce new relation $distance(Edge, Distance)$ where Edge is the edge between the two nodes.}
\label{table:object-relations-all}
\end{table}

\begin{table}[h!]
\centering
\resizebox{\columnwidth}{!}{
\begin{tabular}{lp{0.5\columnwidth}}
\hline
Filter & Description \\ \hline \hline

$filterByColor(Node, Color)$ & return True if Node has Color\\
$filterBySize(Node, Size)$ & return True if Node is of Size \\
$filterByNeighborColor(Node, Color)$ & return True if Node has neighbor with Color\\
$filterByNeighborSize(Node, Size)$ & return True if Node has neighbor with Size\\
\hline
\end{tabular}}
\caption{{\bf Base Filters}.}
\label{table:filtersfull}
\end{table}

\begin{table}[h]
\centering
\resizebox{\columnwidth}{!}{
\begin{tabular}{lp{0.5\columnwidth}}
\hline
Transformation & Description \\ \hline \hline

$updateColor(Node, Color)$ & Update color of Node to Color\\
$move(Node, Direction)$ & Update pixels of Node to move 1 pixel in Direction \\
$moveMax(Node, Direction)$ & Update pixels of Node to move in Direction until it collides with another node \\
$rotate(Node)$ & Update pixels of Node to rotate it clockwise\\
$fillRectangle(Node, Color)$ & Fill background nodes in rectangle enclosed by the node with Color\\
$hollowRectangle(Node, Color)$ & Color all nodes in rectangle enclosed by the node with Color\\
$addBorder(Node, Color)$ & Add additional pixels to Node in Direction\\
$insertPattern(Node, Pattern)$ & Insert Pattern at Node \\
$mirror(Node, Pixel, Direction)$ & Mirror Node toward Direction around Pixel\\
$extend(Node, Direction)$ & Add additional pixels to Node in Direction\\
$flip(Node, Direction)$ & Flip Node in place in some direction\\

\hline
$transform(N, [k])$ & Generic transformation with k parameters.\\
\hline
\end{tabular}
}
\caption{{\bf Full List of Transformations}.}
\label{table:transformationsfull}
\end{table}

\begin{table}[h]
\resizebox{\columnwidth}{!}{
\begin{tabular}{lll}
Actual & Predicted  & Penalty \\ \hline
Background  & Non-background  & 2 \\
Non-background  & Background  & 2 \\
Non-background  & Non-background wrong color  & 1 \\
Non-background  & Non-background right color  & 0 \\
Background  & Background &  0\\
\end{tabular}}
\caption{\bf Heuristic Function used in Search}
\label{table:cost-function}
\end{table}

\begin{table}[htbp!]
\centering
\resizebox{\columnwidth}{!}{
\begin{tabular}{lp{0.5\columnwidth}}
\hline
Constraint & Description \\ \hline \hline
$positionUnchanged (Node, Node)$ & Node does not change position after update\\
$colorUnchanged (Node, Node)$ & Node does not change color after update\\
$sizeUnchanged (Node, Node)$ & Node does not change in size after update\\
\hline
$constraint(Node, Node)$ & Generic constraint\\
\hline
\end{tabular}}
\caption{{\bf Example Constraints}.}
\label{table:constraints}
\end{table}

\begin{table}
\centering
\begin{tabular}{l|lllll}
\hline
     & ARGA                       & -CA                        & -SF                        & -TL                        & -H                         \\ \hline
ARGA & \cellcolor[HTML]{C0C0C0}57 & 54                         & 53                         & 56                         & 56                         \\
-CA  & 54                         & \cellcolor[HTML]{C0C0C0}55 & 51                         & 53                         & 54                         \\
-SF  & 53                         & 51                         & \cellcolor[HTML]{C0C0C0}54 & 53                         & 52                         \\
-TL  & 56                         & 53                         & 53                         & \cellcolor[HTML]{C0C0C0}57 & 55                         \\
-H   & 56                         & 54                         & 52                         & 55                         & \cellcolor[HTML]{C0C0C0}57 \\ \hline
\end{tabular}
\caption{{\bf Solved Tasks Overlaps}.}
\label{table:solvedoverlaps}
\end{table}

\end{document}